%% file: main.tex
\documentclass{article}

\usepackage[english]{babel}

\usepackage[letterpaper,top=2cm,bottom=2cm,left=3cm,right=3cm,marginparwidth=1.75cm]{geometry}

\usepackage{amsmath}
\usepackage{graphicx}
\usepackage[colorlinks=true, allcolors=blue]{hyperref}

\usepackage{graphicx}

\usepackage{tikz}
\usepackage{comment}
\usepackage{amsmath,amssymb} 
\usepackage{color}

\usepackage{hyperref}
\usepackage{bm}
\usepackage{xcolor}
\usepackage[linesnumbered,ruled,vlined]{algorithm2e}
\usepackage{times}
\usepackage{epsfig}
\usepackage{graphicx}
\usepackage{multirow}
\usepackage{verbatim}
\usepackage[linesnumbered,ruled,vlined]{algorithm2e}
\usepackage[acronym,smallcaps]{glossaries}
\usepackage{enumitem}

\title{Visualizing Global Explanations of Point Cloud DNNs}
\author{Hanxiao Tan \\
AI Group, TU Dortmund, Germany \\
hanxiao.tan@tu-dortmund.de}
\date{}
\begin{document}
\input{acronyms}

\newcommand{\beginsupplement}{%
        \setcounter{section}{0}
        \renewcommand{\thesection}{S\arabic{section}}%
        \setcounter{table}{0}
        \renewcommand{\thetable}{S\arabic{table}}%
        \setcounter{figure}{0}
        \renewcommand{\thefigure}{S\arabic{figure}}%
        \setcounter{equation}{0}
        \renewcommand{\theequation}{S\arabic{equation}}%
     }

\maketitle

\begin{abstract}
So far, few researches have targeted the explainability of point cloud neural networks. Part of the explainability methods are not directly applicable to those networks due to the structural specifics. In this work, we show that Activation Maximization (AM) with traditional pixel-wise regularizations fails to generate human-perceptible global explanations for point cloud networks.  We propose generative model-based AM approaches to clearly outline the global explanations and enhance their comprehensibility. Additionally, we propose a composite evaluation metric that simultaneously takes into account activation value, diversity and perceptibility to address the limitations of existing evaluating methods. Extensive experiments demonstrate that our generative-based AM approaches outperform regularization-based ones both qualitatively and quantitatively. To the best of our knowledge, this is the first work investigating global explainability of point cloud networks. Our code is available at: \url{https://github.com/Explain3D/PointCloudAM}.
\end{abstract}

\input{1_Introduction}
\input{2_RelatedWork}
\input{3_Methods}
\input{4_Experiments}
\input{5_Conclusion}

\bibliographystyle{alpha}
\bibliography{egbib}

\end{document}

%% file: acronyms.tex
\newacronym{pc}{PC}{point cloud}
\newacronym{ig}{IG}{Integrated Gradients}
\newacronym{dnn}{DNN}{deep neural networks}
\newacronym{am}{AM}{Activation Maximization}
\newacronym{cnn}{CNN}{convolutional neural networks}
\newacronym{fid}{FID}{Fréchet inception distance}
\newacronym{is}{IS}{Inception Score}
\newacronym{ae}{AE}{AutoEncoder}
\newacronym{aed}{AED}{AutoEncoder with Discriminator}
\newacronym{naed}{NAED}{Noisy AutoEncoder with Discriminator}
\newacronym{pcams}{PC-AMs}{Point Cloud-Activation Maximization Score}
\newacronym{mis}{m-IS}{modified Inception Score}
\newacronym{cd}{CD}{Chamfer Distance}
\newacronym{emd}{EMD}{Earth Mover's Distance}

%% file: 1_Introduction.tex
\section{Introduction}
Point clouds are one of the most widely used data forms of 3D representation and have been extensively used for various applications, such as autonomous driving and robotics. Due to their disordered property, traditional CNNs are not directly applicable to point cloud data. Recently, several studies have proposed multifarious deep learning approaches for point clouds \cite{qi2017pointnet,qi2017pointnet++,wang2019dynamic} that achieved state-of-the-art accuracies in existing benchmark datasets. So far, however, very little attention has been paid to the trustworthiness of point cloud networks. Since such black-models are widely used - even in high-risk applications - while generally struggle for inspiring trust in users and remain almost impossible to debug, this research gap needs to be addressed.

The research of explainability plays an important role in addressing the issue of trustworthy AI. Previous studies proposed a considerable number of explainability approaches including gradient-based \cite{simonyan2014deep,Bach2015,shrikumar2019learning,springenberg2015striving,smilkov2017smoothgrad,Sundararajan2017} and local surrogate model-based \cite{Ribeiro2016,Lundberg2017,ribeiro2018anchors}, which generate post-hoc local explanations to a specific input instance. Global explainability approaches are another series that allow for an inclusive explanation of the entire black-box model, such as surrogate model simplification \cite{kamruzzaman2010algorithm} and \gls{am} \cite{Nguyen2019}. Although the aforementioned approaches facilitate the faithfulness of models dealing with tabular and image data, there has been little discussion about the explainability of point cloud networks. Due to the specific architecture, point cloud networks possess distinctive properties from traditional multi-width convolutional neural networks (for instance, \cite{gupta20203d} found the features learned by point cloud networks are extremely sparse), suggesting that explainability studies on point cloud networks may lead to novel discoveries.

On the other hand, it is difficult to quantitatively evaluate the accuracies of the generated explanations due to the lack of ground truth. Human evaluations are highly subjective and therefore lack persuasiveness and reproducibility. For \gls{am}, several previous studies have used quantitative metrics to evaluate the quality of the synthesized images \cite{nguyen2017plug,mishra2019gan,zhou2017activation}. However, we argue that the performance assessed by these traditional metrics is neither comprehensive nor can be deceived by AM of point cloud networks.

This work strives to investigate the global explanations of the popular point cloud networks with \gls{am}. We show that non-generative network-based \gls{am} approaches for images are not applicable to point clouds (see figure \ref{fig:2Dvanilla}), and propose generative \gls{am} methods for the global explainability of point cloud networks. Additionally, we propose a more persuasive and comprehensive evaluation metric for point cloud \gls{am}, and demonstrate that our point cloud \gls{am} methods outperform all other methods both at the human cognitive level and in quantitative assessment. Our contributions are primarily summarized as follows:

\begin{itemize}
    \item As the first work investigating global explainability of point cloud networks, we exhibit that non-generative \gls{am} methods are unable to generate human-comprehensible explanations. Addressing the challenge, we propose generative model-based \gls{am} approaches that depict the global peculiarities of point cloud networks.
    \item We propose a convincing evaluation metric for point cloud \gls{am}: \gls{pcams}, which simultaneously captures the activation value, diversity, human perception-level and physical-level authenticity of generated \gls{am} examples.
\end{itemize}

The rest of this paper is organized as follows: Section \ref{related work} introduces explainability methods for point clouds, especially \gls{am} and corresponding evaluation methods. Section \ref{methods} provides the proposed generative \gls{am} approaches for point clouds as well as a more persuasive evaluation metric. Section \ref{experiments} demonstrates our experimental results and we summarize our work in section \ref{conclusion}.

%% file: 2_RelatedWork.tex
\section{Related Work} \label{related work}
In this section, we introduce the popular explainability methods, review the proposed \gls{am} approaches, and state the current progress of explainability research on point cloud neural networks.

\textbf{Explainability methods}: In contrast to interpretability approaches that render the decision process understandable, explainability methods aim to elucidate the operating principles of black-box models with mechanisms that are asynchronous with the decision-making periods. Explainability methods are categorized into two groups according to their objects: local and global explainers.

\textit{Local explainers} typically generate explanations corresponding to individual inputs by tracing gradients \cite{simonyan2014deep,Bach2015,shrikumar2019learning,springenberg2015striving,smilkov2017smoothgrad,Sundararajan2017} or employing surrogate models and perturbations \cite{Ribeiro2016,Lundberg2017,ribeiro2018anchors}. Nonetheless, gradient-based explainability methods are considered noisy, and in recent sanity studies, part of the methods were found to be model-independent \cite{adebayo2018sanity}. Surrogate model-based approaches require extensive perturbation instances as training datasets and are therefore computationally intensive. Another common drawback of local explainability methods is the lack of holistic views of the overall datasets, compounding the cost of intrinsically understanding the decision process.

\textit{Global explainers} provide explanations in regard to entire datasets rather than individual input instances by demonstrating its inherent characteristics. The global explanation may not be precise for each classification case, however, it provides a more intuitive representation of how the model works. Global explanations are typically presented in the following forms: \cite{kamruzzaman2010algorithm} extracts decision rules from the original model that is comprehensible for users, \cite{casalicchio2018visualizing,covert2020understanding} rank the aggregated feature importance according to the whole datasets. For computer vision tasks, listing the feature importance is challenging because of the extensive number of unaligned features. As an alternative, \gls{am} is thus proposed to exhibit intuitive global explanations by generating highly representative examples of specific classes.

\textbf{Activation Maximization (AM)}: \gls{am} is a high-level feature visualization technique that was first proposed by \cite{erhan2009visualizing}. AM chooses a target activation unit and maximizes it by optimizing the input vector while freezing all other neurons in the DNN. However, without incorporating any prior or constraints, \gls{am} will synthesize mosaic images that are incomprehensible to humans and are not explainable \cite{nguyen2015deep}. Optimization constrains, such as L2-norm \cite{simonyan2013deep}, Gaussian blur \cite{yosinski2015understanding}, Total Variation \cite{mahendran2016visualizing} or priors, such as average image initialization \cite{nguyen2016multifaceted} and patch dataset \cite{wei2015understanding,mordvintsev2015inceptionism}, successfully synthesize object images with clear outlines, and therefore facilitate the explainability. Another solution for enhancing the comprehensibility of \gls{am} images is to learn the distribution of real objects with generative models. \cite{nguyen2016synthesizing,zhou2017activation,mishra2019gan,katzmann2021explaining} utilized auto-encoders and GANs to produce high quality \gls{am} images. \cite{nguyen2017plug} proposed Plug \& Play embedding generative networks that simultaneously address the high-resolution and diversity of synthesized \gls{am} images. Additionally, \cite{xiao2019gradient} proposed a black-box \gls{am} approach based on evolutionary algorithms. Nevertheless, point clouds are structurally different from traditional image \gls{dnn}s so that the aforementioned AM methods are not directly applicable to point cloud networks.

On the other hand, evaluating the quality of AM images is challenging and so far, most previous works rely on subjective human intuition as the evaluation criterion. \cite{nguyen2017plug} accessed the definition and diversity of AM images via \gls{is} \cite{salimans2016improved}. \cite{mishra2019gan} incorporated \gls{fid} \cite{heusel2017gans} to estimate the similarity between generated AM examples and real instances in latent spaces. AM score, another evaluation metric proposed by \cite{zhou2017activation}, is ameliorated from \gls{is} and addresses the uneven distribution of data categories.

\textbf{Explainability research on point clouds}: There are relatively few explainability studies in the area of point clouds. \cite{Zheng_2019_ICCV} traced the critical points to generate saliency maps of the point cloud network by dropping points. \cite{gupta20203d} was the first work to incorporate explainability methods, who started an observation of the intrinsic feature of point cloud networks via \gls{ig}. A follow-on study was conducted by \cite{tan2022surrogate}, which proposed a local surrogate model-based approach for explaining point cloud networks. However, one limitation of the approaches mentioned above is that local explainability methods are only concerned with specific inputs that can hardly present the intrinsic properties of the whole point cloud network.

%% file: 3_Methods.tex
\section{Methods} \label{methods}
In this section, we demonstrate our \gls{am} approach for point clouds (section \ref{mtd:3DAM}) as well as the proposed evaluation metric for point cloud \gls{am} (section \ref{mtd:eva}).
\subsection{3D Activation maximization} \label{mtd:3DAM}
\begin{figure*}
\centering
\includegraphics[width=1.0\textwidth]{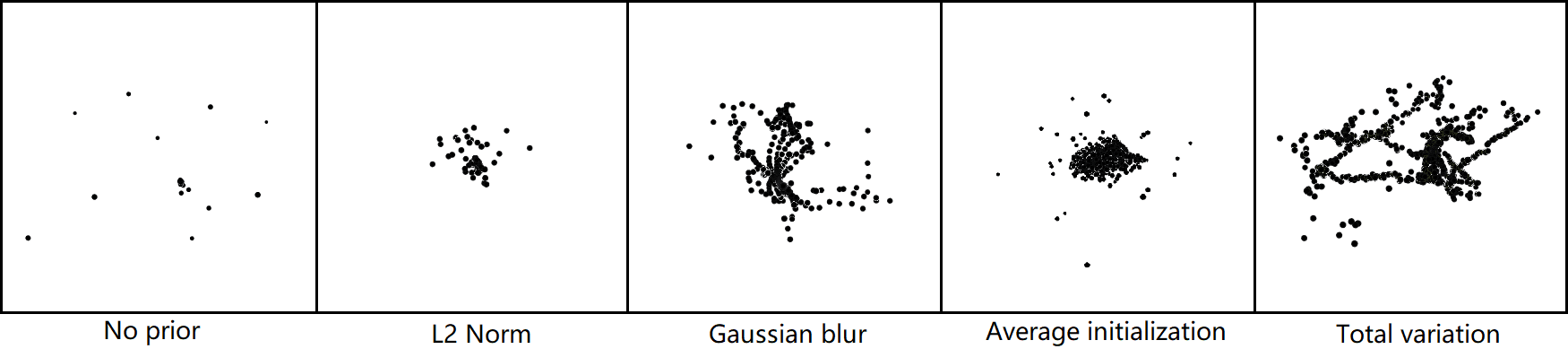}
\caption{\gls{am} for point clouds without generative priors. Due to the specific architecture of the point cloud network, traditional regularization priors (for 2D images) are incapable of generating human-perceivable global explanations.}
\label{fig:2Dvanilla}
\end{figure*}

To visualize a specific neuron in the DNN, \cite{erhan2009visualizing} proposed the \gls{am}, which is formulated as:
\begin{equation} \label{eq:am}
x^* = \underset{x}{\mathrm{argmax}}\, (a_i^l(\theta,x))
\end{equation}
where $x$ and $\theta$ denote the input instance and the parameters in the DNN respectively, and $a_i^l(\theta,x)$ denotes the $i^{th}$ neuron at $l^{th}$ layer. From an explainability perspective, setting a neuron in the highest layer of the network as the target activation, \gls{am} can be considered as providing a global explanation of the entire network, i.e., an ideal input for a particular class \cite{samek2019explainable}. However, 2D \gls{am} without any prior suffers from generating examples with high-frequency mosaics that are unrecognizable to humans \cite{nguyen2015deep}. Several studies have investigated regularizing AM examples with non-generative priors, such as L2 Norm, Gaussian blur and Total variation \cite{simonyan2013deep,yosinski2015understanding,mahendran2016visualizing}. While the above mentioned enhancements have made progress in human interpretability for 2D images, their effectiveness is severely compromised while processing point clouds (see figure \ref{fig:2Dvanilla}). We believe that on the one hand, the features of point cloud networks are comparatively sparse and the global structure information of instances is seriously impaired \cite{gupta20203d}, and on the other hand, the adjacency-based regularizations fail due to the disorderliness of point clouds.

To address the scarcity of structural information, we propose generative model-based AM to interpret the global properties of point cloud networks. The outputs $O_{g}$ to be searched are subject to two obligatory restrictions simultaneously: they highly activate a neuron at a high level of the networks (equation \ref{eq:am}) and are under the similar distribution as the dataset that is recognizable for humans $O_{g} \sim X$. Currently \gls{am}s are only accomplished via gradient-based optimizers, therefore, we first establish a function that generates point clouds with realistic distributions, then we filter out those samples that highly activate neurons from the output of the function by \gls{am}.

\begin{figure*}
\centering
\includegraphics[width=1.0\textwidth]{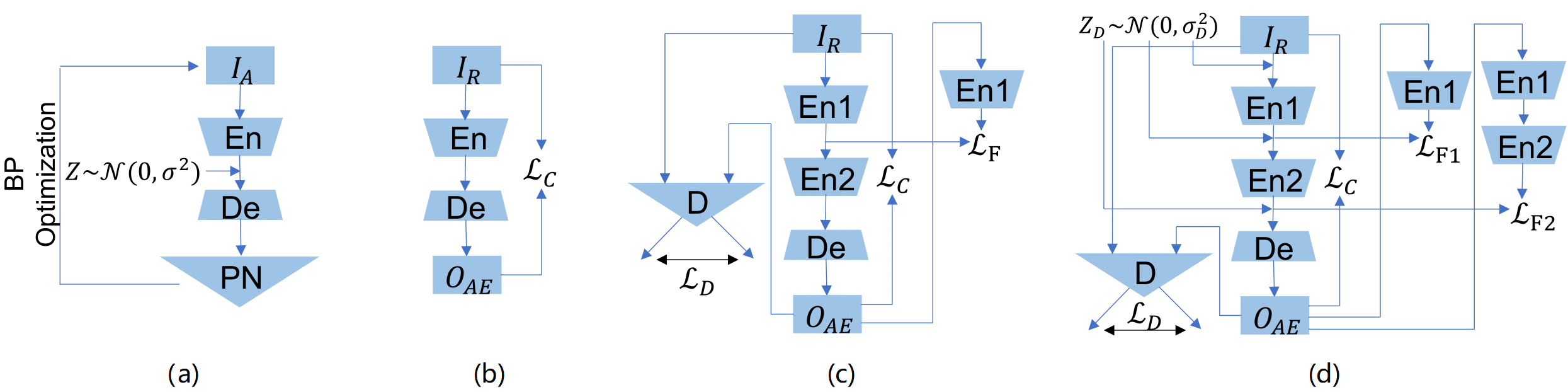}
\caption{Generative \gls{am} architectures for point clouds. (a) General \gls{am} optimization structure. (b) \gls{am} based on \gls{ae}. (c) \gls{am} based on \gls{aed}. (d) \gls{am} based on \gls{naed}. $En$, $De$, $D$ and $PN$ denote encoders, decoders, discriminators and the PointNet structure respectively, and $I_A$, $I_R$, $O_{AE}$ represent the averages of the test dataset, real instances and the output of the \gls{ae} respectively.}
\label{fig:3DGenmodel}
\end{figure*}

\textbf{\gls{ae}}: The most intuitive approach is to learn the point distribution via \gls{ae} ((a) in \ref{fig:3DGenmodel}). \cite{achlioptas2018learning} has demonstrated that a symmetric pooling layer followed by point-wise convolutions is capable of encoding point clouds into 1-dimensional latent representations, while a simple multi-layer, fully connected network is qualified as the decoder. Following the structure, we first train the \gls{ae} with \gls{cd} loss $\mathcal{L}_C$. Our \gls{am} architecture based on \gls{ae} is presented in (b) of figure \ref{fig:3DGenmodel}. During optimization, we observe that the optimizer is frequently trapped in local optimum and struggles to further activate the target neuron. We believe the reason to be that the random initialized point distribution is excessively divergent from the reality and propose two solutions. Inspired by \cite{nguyen2016multifaceted}, we encode the average of the dataset as the initialization of the input latent vector, and add Gaussian noise if the optimization process is stuck.

\textbf{\gls{aed}}: One flaw of the pure \gls{ae} is the diversity of the generated samples. Instead of forcing the \gls{ae} into multifarious \gls{am} outputs, adding random noise without any restriction is inclined to downgrade the quality of the output. Addressing this problem, we introduce a discriminator, which acts similarly to GANs. \gls{ae} tries to fool the discriminator by generating fake instances that mislead the discriminator to classify them as true objects, while the discriminator attempts to correctly identify both. Our discriminator shares the same structure as the T-net in PointNet \cite{qi2017pointnet}, except for replacing the fully connected layer at the highest level with a Sigmoid function. Inspired by \cite{nguyen2017plug}, in addition to the point-wise \gls{cd}, we incorporate another latent vector distance loss $\mathcal{L}_F$, which measures the feature distinction between two instances after being encoded by the $1 \times 1$ convolutional layer of PointNet. The final generative loss is formulated as:
\begin{equation}
    \mathcal{L}_g = \mathcal{L}_C + w_F\mathcal{L}_F - w_D\mathcal{L}_{D_{f}}
\end{equation}
where $w_F$, $w_D$ are the corresponding weights and $\mathcal{L}_{D_{f}}$ denotes the loss of discriminator to distinguish fake instances. The overview of the architecture is in (c) in \ref{fig:3DGenmodel}. In the training process, we observe that since the performance of the discriminator easily outperforms the \gls{ae} ($\mathcal{L}_D \ll 0$) \cite{arjovsky2017wasserstein}, the latter struggles to be further optimized. We therefore train only one of them alternately for each batch: If $\mathcal{L}_D < 0$, we train the \gls{ae} only and vice versa. Furthermore, if the discriminator is overperforming ($\mathcal{L}_D < -0.75$), we add Gaussian noise to its parameters to disrupt the performance.

\textbf{\gls{naed}}: To further reinforce the diversity and robustness, we attempt to introduce Gaussian noise to the parameters of the \gls{ae} while training. For regularization, another latent feature distance is incorporated under the second encoder. A primary component of the second encoder is max-pooling, which extracts the global features of the instance. Besides the point-wise feature loss, another loss $\mathcal{L}_{F2}$ is attached to measure the distance on the global level. The remaining configurations are identical to those in \gls{aed}. The final loss is formulated as:
\begin{equation}
    \mathcal{L}_g = (\mathcal{L}_C + w_{F1}\mathcal{L}_{F1} + w_{F2}\mathcal{L}_{F2}) - w_D\mathcal{L}_{D_{f}}
\end{equation}

\subsection{Evaluation metrics for Point Clouds AM} \label{mtd:eva}
For the evaluation of explainability methods, most previous research evaluates them by showing examples to humans. However, subjective assessments are avoided whenever possible. A fair evaluation method should not only quantitatively assess the generated results computationally, but also be consistent with human perception, thus promoting the explainability. The existing evaluation metrics, that are applicable to \gls{am}s, can be categorized into three classes:

\textbf{Non-perceptibility metrics}, represented by \gls{is} \cite{salimans2016improved} or AM Score \cite{zhou2017activation}, aim to assess the quality and diversity of \gls{am} generations. However, this series of approaches evaluates the generation quality by calculating the entropy of the logits, while the disparity in human perception levels is absent. For point clouds, they fail to distinguish between AM methods without a prior and those based on generative models, although the latter are apparently more comprehensible to humans {see figure \ref{fig:AMall}}.

\textbf{Pixel-wise metircs}, represented by $L_p$ (2D), Chamfer and Hausdorff distances (3D), address forcing the generated instances to be pixel-wisely approximated to the real objects. Nevertheless, instances that comply with this metric may lose the ability to be "global explainable" as it does not require the instances to be globally representative. Suppose a generator that perfectly reconstructs the original instance, even though the distance loss can be minimized to $0$, but does not facilitate human understanding of the model peculiarities.

\textbf{Latent feature metrics}, represented by \gls{fid}, measure the distinction on the feature level, which are theoretically promising, and are widely applied in 2D generative models. We follow the \gls{fid} from \cite{sun2020pointgrow} which compared the global feature from the PointNet architecture. Nonetheless, we observe that the metric is vulnerable for \gls{am} (see table \ref{table:methodcompare}: randomly initialized instances achieve \gls{fid} scores as high as those from generative models, though they are not perceived by humans). We believe that the \gls{fid} is affected to some extent by the sparsity of the point clouds due to the scarcity of adjacent relations in the point cloud networks.

\textbf{\gls{pcams}}: Targeting the limitations of the aforementioned methods, we propose a composite \gls{am} evaluation metric: \gls{pcams}. Our \gls{pcams} is formulated as:
\begin{equation}
    PCAMS = IS_m - \frac{(log(FID_{PN}) + log(CD))}{2}
\end{equation}
$IS_m$ denotes the \gls{mis} \cite{gurumurthy2017deligan}, which is formulated as: 
\begin{equation}
    IS_m=e^{{\mathbb{E}_{x_i}}[\mathbb{E}_{x_j}[(\mathbb{KL}(p(y|x_i)||p(y|x_j))]]}
\end{equation}
where $x_i$ and $x_j$ denotes different instances with the same label. In addition to the values of the corresponding activations, \gls{mis} concentrates more on the diversity of the generated examples within classes than the variety of inter-class labels. Therefore we utilize the \gls{mis} which employs the cross-entropy of the predictions within intra-class examples.

$FID_{PN}$ denotes the PointNet-based \gls{fid} and is formulated as:
\begin{equation}
    FID_{PN}=\left \| \mu_r -\mu_g  \right \|^{2}+Tr(\sigma_r + \sigma_g - 2(\sigma_r \sigma_g)^{\frac{1}{2}})
\end{equation}
where $A_r \sim \mathcal{N}(\mu_r,\sigma_r)$ and $A_g \sim \mathcal{N}(\mu_g,\sigma_g)$ are the activations from the reference network, which are approximately considered as Gaussian distributions. \gls{fid} measures the distance between the two distributions, lower \gls{fid} scores imply closer proximity of the generated examples to the real instances, and therefore higher perceptibility. Nevertheless, the standard reference network \textit{Inception-v3} is no longer applicable to $FID_{PN}$ since the multi-width convolutional kernel for images fails to extract adjacent features from unordered point clouds. Following \cite{sun2020pointgrow}, we substitute the backbone of PointNet for Inception-v3 and choose from the layers above the max-pooling (global features) as the activation.

Due to the fragility of $FID_{PN}$, we introduce an additional perceptibility measure: \gls{cd}, formulated as:
\begin{equation}
    CD(x_g,x_i)=\frac{1}{|x_g|}\sum_{p_m\in x_g}\min_{p_n\in x_i}\left \| p_m-p_n \right \|_2
\end{equation}
Although \gls{cd} estimates the similarity between examples more precisely, it lacks generality as a scoring criterion for \gls{am}. To alleviate this deficiency, we randomly draw several instances from the dataset with the same labels as the generated examples and calculate the average of the \gls{cd}s. We finalize the aforementioned three metrics by logarithmically scaling \gls{fid} and \gls{cd} to the same order of magnitude with \gls{mis}, such that the final score does not collapse due to the numerical explosion of any single term.

%% file: 4_Experiments.tex
\section{Experiments} \label{experiments}
In this section, we qualitatively demonstrate the generated examples of our proposed point cloud-applicable \gls{am} (section \ref{exp:visu}), and show the quantitative evaluations of existing point cloud \gls{am} approaches (section \ref{exp:eva}). Additionally, we also provide an example of application scenes of proposed methods for prediction examination in section \ref{sup: datareview}. In our experiments, we choose ModelNet40 \cite{Wu2015} as our dataset, which contains $12311$ CAD models in $40$ common classes and is currently the most widely-used point cloud dataset. Besides, we also test our approaches on the classification set of ShapeNet \cite{chang2015shapenet}, which is composed of $45969$ point cloud instances ($35708$ for training and $10261$ for testing) in $55$ classes. We select PointNet as our primary experimental model, which is the pioneer of deep learning for raw point clouds. We also validate our result in the most popular point cloud models i.e., PointNet++ \cite{qi2017pointnet++} and DGCNN \cite{wang2019dynamic}. During \gls{am} generation, we heuristically set the latent dimensions of the \gls{ae} as $128$ and the learning rate as $5e-6$. The \gls{am} optimization stops after $2 \times 10^{4}$ iterations. For quantitative evaluation, we generate $10$ \gls{am} examples for each class, and we randomly select $5$ real instances from the dataset as the baseline for calculating \gls{fid}s and \gls{cd}s and average the corresponding results.

\subsection{Point Cloud AM Visualization} \label{exp:visu}
\begin{figure*}
\centering
\includegraphics[width=1\textwidth]{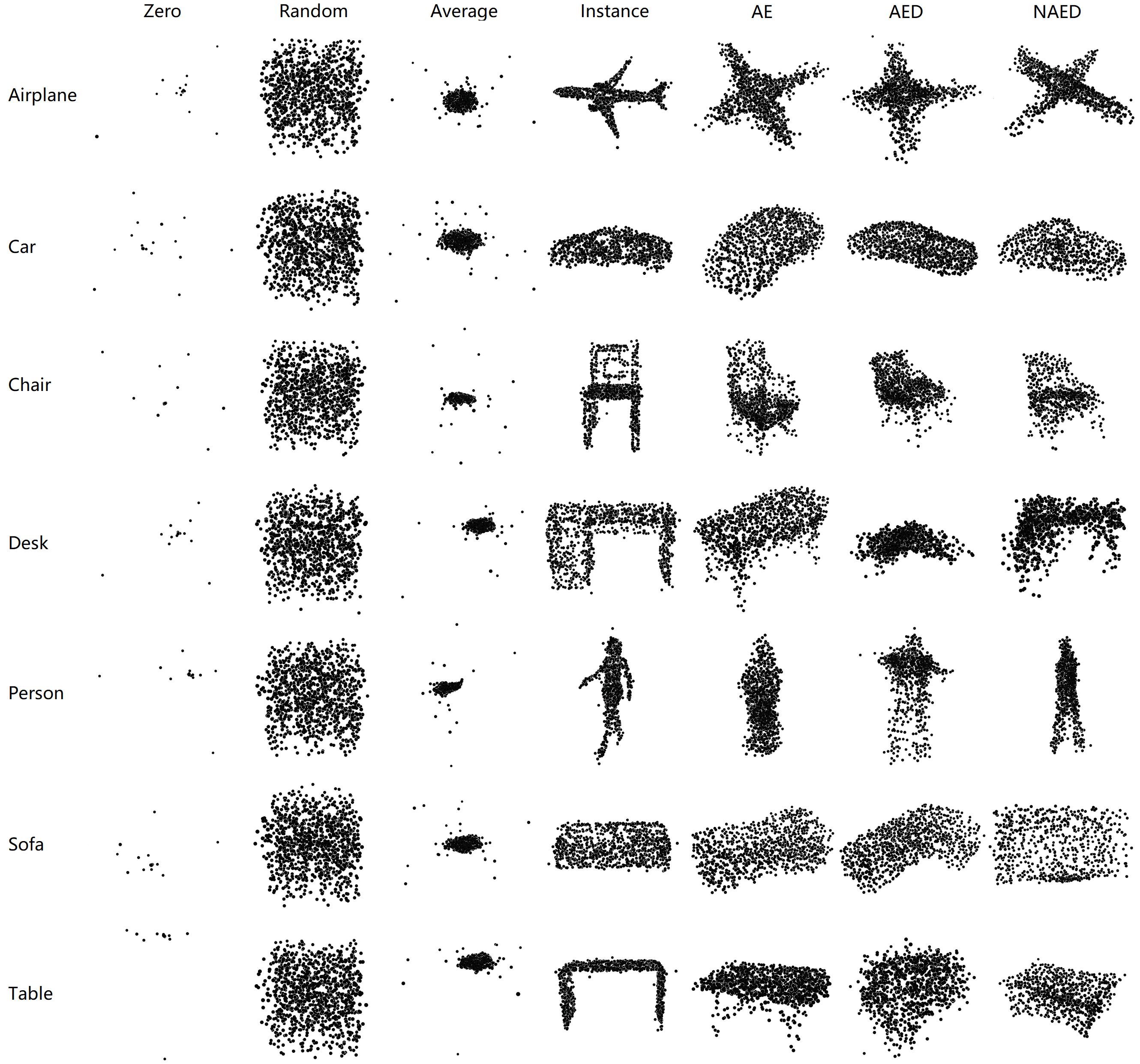}
\caption{\gls{am} results of different approaches. From left to right: Zero initialization, random initialization, initialized with the average of the test data per class, initialized from a specific instance, and our proposed \gls{ae}, \gls{aed} and noisy \gls{naed}}.
\label{fig:AMall}
\end{figure*}

\textbf{Perceptibly}: Figure \ref{fig:AMall} shows the point cloud \gls{am} examples of common classes generated by multifarious approaches on ModelNet40. Zero and random initialization, while highly activating the selected neurons, results in only the expansion of individual points due to the lack of a prior and therefore fails to yield human understandable global explanations. Initialization with the average of the test data performs better in 2D images. However in point clouds, explainability is not significantly enhanced compared to the no-prior methods since the point cluster in the center struggles to render the distribution of common objects. Initialized from a specific instance though outlines the objects best, nevertheless, the information of the "global" is absent i.e., the general distribution of the whole dataset. The contours of the objects are derived from the input instances themselves rather than the global activation-optimization process. The former tends to expose more local information about particular inputs and is therefore more generally utilized in adversarial attacks. Note that due to the properties of point cloud networks, the aforementioned non-prior or point prior methods prefer to optimize individual points rather than adjusting the global distribution, which leads to surface discontinuities in the output 
examples. In comparison, our generators with latent priors dominate in terms of both shape consistency and human perceptibly.

Among the generative methods, \gls{am} examples provided by \gls{ae} are intuitively more stable, especially compared to those from \gls{aed}. We believe this is due to the absence of noise mechanisms and the singularity of the loss term. In \gls{ae}, no noise is incorporated except for the neuron maximization module that prevents the optimization process from sticking in local optimums, and the generator is trained via an one-fold \gls{cd} loss which forces the output to be point-wise approximated to real objects. These mechanisms regularize the profile of the generated examples to be reconstructed precisely as the real instances from the dataset while the outputs suffer from a scarcity of distinction. On the other hand, in \gls{aed} and \gls{naed}, the multi-fold loss functions balance the constraints of approximating the dataset in both point-wise and latent feature levels. Compared to \gls{ae}, this module causes a few collapses of the output geometries, but by introducing adversarial learning with a discriminator, the generator is still able to reconstruct the contours of real objects and enrich their diversity simultaneously. Moreover, we surprisingly find that incorporating cascaded Gaussian noise to the encoder during training further enhances the quality and diversity of the \gls{am} outputs. We present the generation diversity in the next subsection.

\begin{figure}
    \centering
    \includegraphics[width=0.8\textwidth]{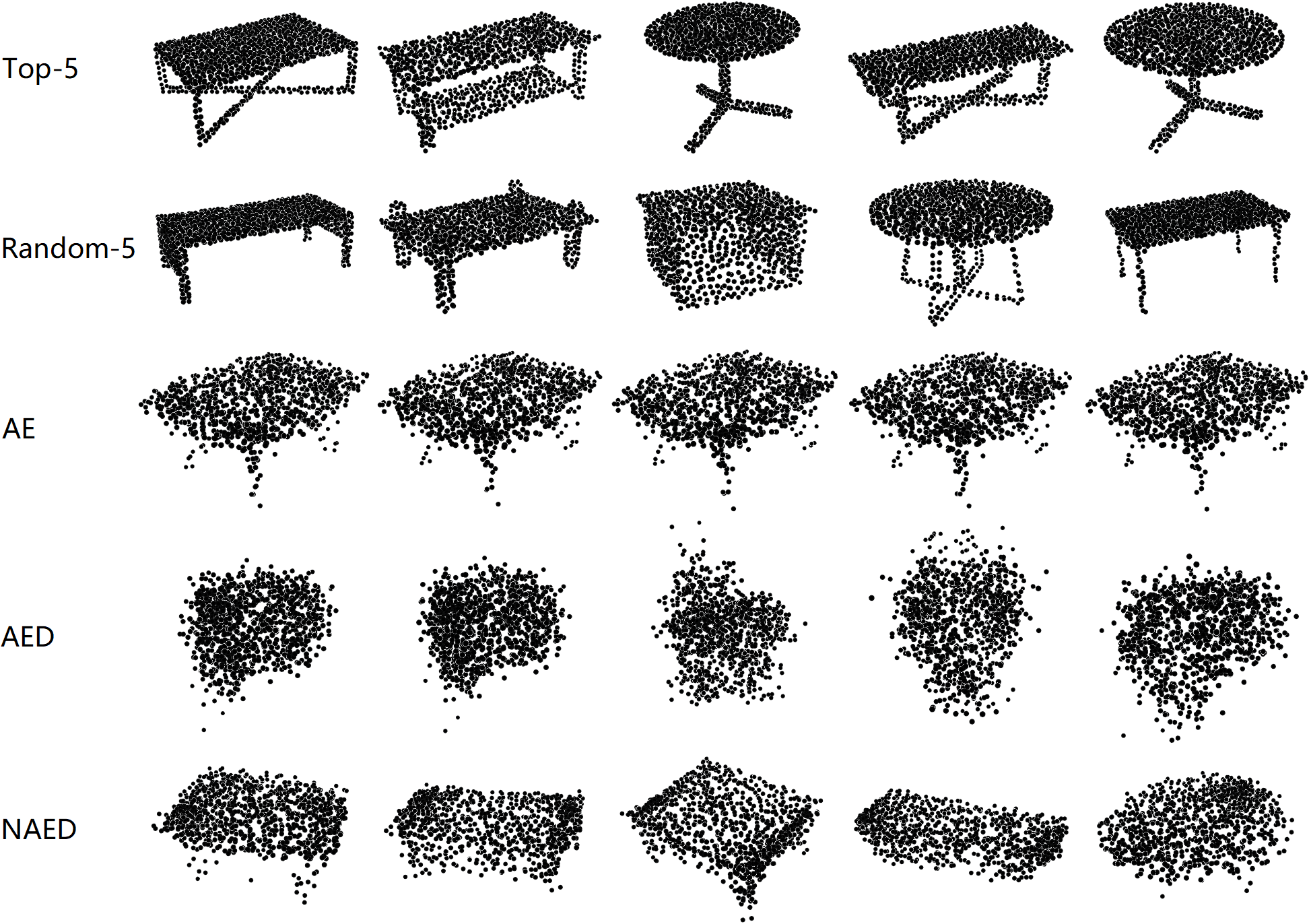}
    \caption{Diversity of \gls{am} generations. We choose $5$ examples from instances that 1) most hightly activate the neuron 2) are selected randomly 3) are from the generations of \gls{ae} 4) of \gls{aed} 5) of \gls{naed}.}
    \label{fig:Diversity}
\end{figure}

\textbf{Diversity}:
Another key point of \gls{am} quality is the diversity. In figure \ref{fig:Diversity}, we visualize $5$ examples for each one of the generative \gls{am} methods which are randomly selected from the generation repository. We also demonstrate the five examples in the dataset that most highly activate the neuron "table", as well as five stochastically selected examples respectively for references. As can be seen from the figure, \gls{ae} is more stable than the others, while lacks diversity. In comparison, both \gls{aed} and \gls{naed} depict the multiplicity of the objects while \gls{aed} is somewhat deficient in terms of stability.

\begin{figure}
    \centering
    \includegraphics[width=0.5\textwidth]{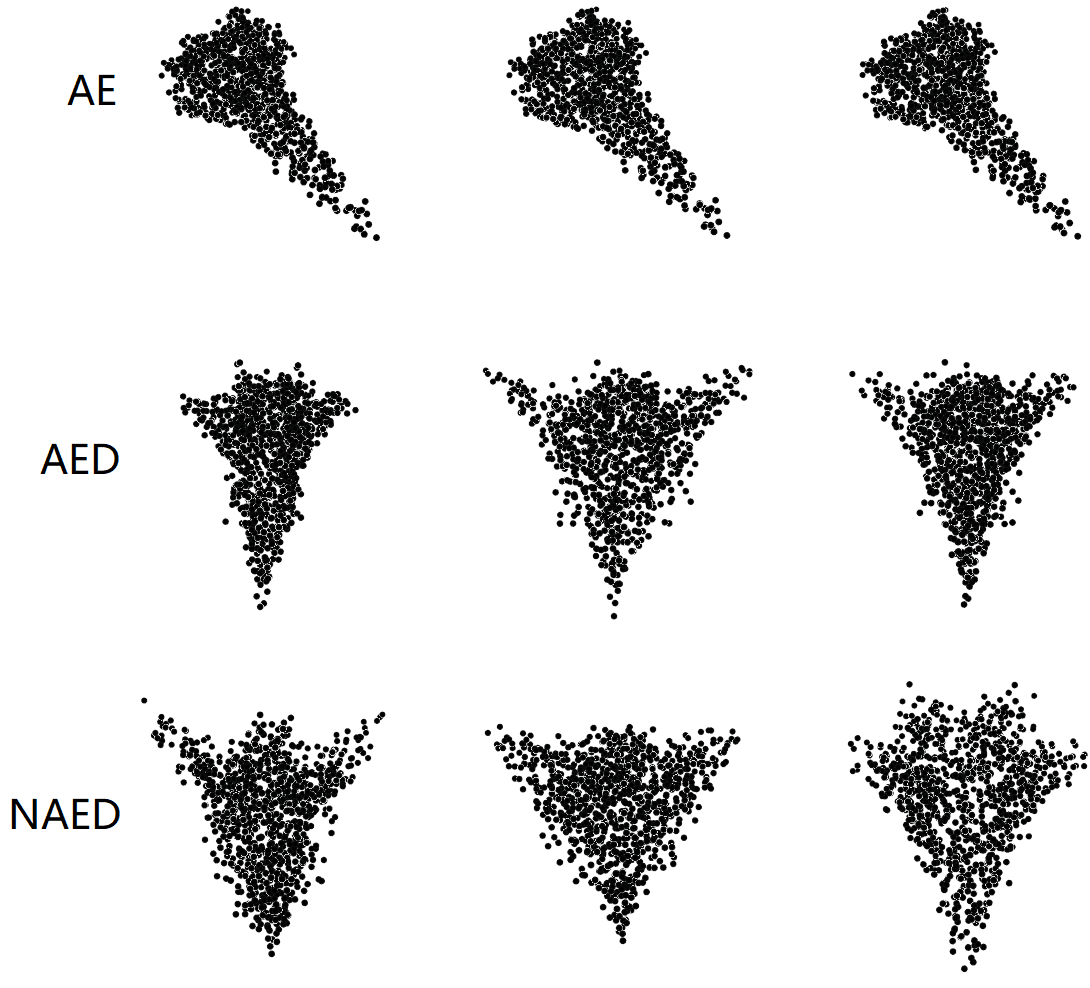}
    \caption{\gls{am} examples from \gls{ae}, \gls{aed} and \gls{naed} respectively of class "airplane" in ShapeNet.}
    \label{fig:Shapenet}
\end{figure}

\textbf{Experiments on ShapeNet}: We also present the \gls{am} results of the class "airplane" generated by the proposed methods employing ShapeNet as the experimental dataset in figure \ref{fig:Shapenet}. Similar to ModelNet40, the global explanations presented by \gls{ae} also exhibit only minimal spatial offsets, while \gls{aed} and \gls{naed} outperform \gls{ae} in terms of the diversity of object outlines. Subjectively, the samples generated by \gls{naed} are more stable due to the noise introduction in the training process.  

\subsection{Evaluation Metric of Point Cloud AM} \label{exp:eva}

Visually assessing the \gls{am} global explanation is highly subjective, and therefore we quantitatively evaluate the results via the proposed methods in table \ref{table:methodcompare}. Since there is no existing \gls{am} study for point clouds, we consider the \textit{no prior} and \textit{point-wise} prior approaches as our baseline. Note that in terms of \gls{fid}, \gls{am}s with random initialization also achieve a satisfactory loss while the examples are almost indistinguishable by humans, which results in the inability to accurately capture the perceptual distance between samples. Therefore, we introduce \gls{cd} as another regularization. We also incorporate \gls{emd} to validate the approximation of the examples. According to the comparisons, our generative \gls{am} approaches (latent prior) dominate the rest regarding the \gls{pcams}. Though \gls{ae} possesses the minimum distance loss, it suffers from a significant drawback of diversity, which leads to the \gls{mis} being lower than the other approaches ( which is consistent with the demonstrations in figure \ref{fig:Diversity}). In addition, Figure \ref{table:eva on ShapeNet} reports the corresponding evaluations on ShapeNet, where it can be seen that our proposed approaches consistently achieve similar performance on different datasets.

\begin{figure}
    \centering
    \includegraphics[width=0.6\textwidth]{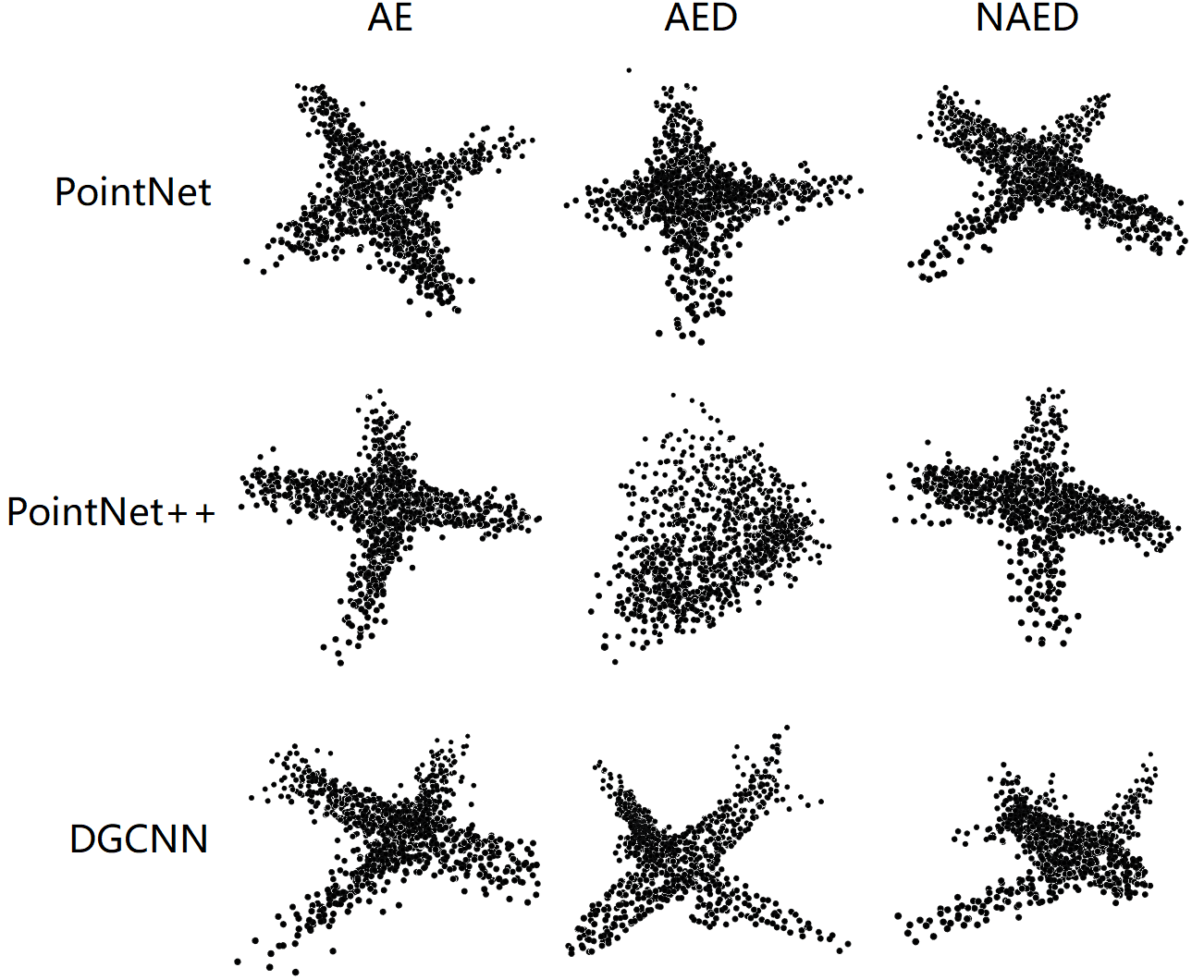}
    \caption{\gls{am} visualization for the most popular point cloud networks: PointNet, PointNet++ and DGCNN.}
    \label{fig:AM PC networks}
\end{figure}

We also evaluate the performance of the proposed \gls{am} methods, applied to different point cloud networks with \gls{pcams}, and present the results in table \ref{table:allmodel}. As a reference, we show a sample of the corresponding visualization in figure \ref{fig:AM PC networks}. Our \gls{ae}-based \gls{am} approaches achieve almost stable performances regarding \gls{pcams}, while being acceptable on other metrics, such as \gls{emd}. We also show the quantitative results of the experiments on ShapNet in the next section.

\begin{table*}[] 
\centering
\begin{tabular}{ccccccc}
\hline
                                  &               & m-IS           & FID            & CD             & EMD             & PC-AMs        \\ \hline
\multirow{2}{*}{No prior}         & Zero          & 1.113          & 0.119          & 0.266          & 364.35          & 2.84          \\
                                  & Random        & 1.081          & 0.016 & 0.245          & 413.52          & 3.85          \\ \hline
\multirow{2}{*}{Point-wise prior} & Average       & 1.001          & 0.097          & 0.230          & 377.20          & 2.90          \\
                                  & Instance      & 1.015          & 0.071          & 0.085          & 228.87          & 3.57          \\ \hline
\multirow{3}{*}{Latent prior}     & AE           & 1.085          & 0.016          & \textbf{0.044} & \textbf{143.13} & 4.71          \\
                                  & AED       & 1.124          & 0.018          & 0.086          & 241.35          & 4.37          \\
                                  & NAED & \textbf{1.461} & \textbf{0.014}          & 0.074          & 207.65          & \textbf{4.89} \\ \hline
\end{tabular}
\caption{\Gls{pcams} evaluation metric for point cloud \gls{am}s. \gls{emd} is also introduced for point-wise distance validation.}
\label{table:methodcompare}
\end{table*}

\begin{table}[]
\centering
\resizebox{0.475\textwidth}{19mm}{
\begin{tabular}{ccccccc}
\hline
                      &       & m-IS           & FID            & CD             & EMD             & PC-AMs        \\ \hline
\multirow{3}{*}{AE}   & PN    & 1.085          & 0.016          & 0.044          & 143.13          & 4.71          \\
                      & PN++   & 1.103          & \textbf{0.008} & \textbf{0.041} & \textbf{134.16} & 5.12 \\
                      & DGCNN & 1.020          & 0.010          & 0.105          & 252.82          & 4.43          \\ \hline
\multirow{3}{*}{AED}  & PN    & 1.124          & 0.018          & 0.086          & 241.35          & 4.37          \\
                      & PN++   & 1.107          & 0.020          & 0.122          & 255.46          & 4.12          \\
                      & DGCNN & 1.358          & 0.013          & 0.109          & 343.15          & 4.63          \\ \hline
\multirow{3}{*}{NAED} & PN    & 1.578  & 0.018          & 0.071          & 353.10          & 4.92          \\
                      & PN++   & \textbf{1.866}          & 0.011          & 0.072          & 236.42          & \textbf{5.43}          \\
                      & DGCNN & 1.316          & 0.015          & 0.109          & 335.51          & 4.52          \\ \hline
\end{tabular}}
\caption{\gls{pcams} evaluations for different point cloud models, where PN and PN++ denotes PointNet and PointNet++.}
\label{table:allmodel}
\end{table}

\begin{table}[]
\centering
\begin{tabular}{cccccc}
\hline
     & m-IS           & FID            & CD             & EMD             & PC-AMs        \\ \hline
AE   & 1.012          & 0.017          & \textbf{0.047} & \textbf{147.87} & 4.57          \\
AED  & 1.146          & 0.012          & 0.076          & 208.02          & 4.65          \\
NAED & \textbf{1.157} & \textbf{0.011} & 0.067          & 203.74          & \textbf{4.75} \\ \hline
\end{tabular}
\caption{Quantitative evaluations on ShapeNet.}
\label{table:eva on ShapeNet}
\end{table}

\begin{figure*}
    \centering
    \includegraphics[width=0.8\textwidth]{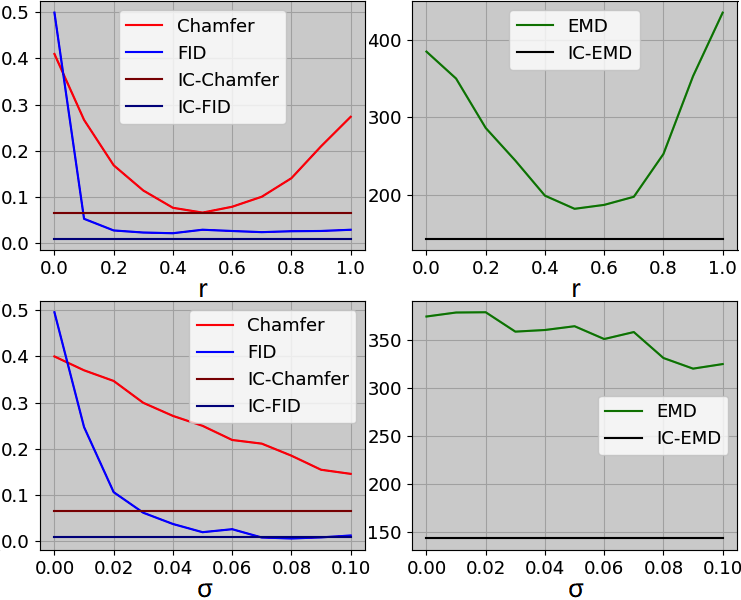}
    \caption{\gls{cd}, \gls{fid} and \gls{emd} metrics of instances generated by various parameters (the first two plots and the last two plots are uniform and Gaussian distributions respectively), where IC denotes intra-class, which is the average of corresponding distances between \textbf{real} objects in the dataset. $r$ denotes the interval parameter of uniform distribution and $\sigma$ denotes the variance of Gaussian distribution. The mean of Gaussian distribution is set to zero to ensure that the generated instances are symmetric about the zero point of the spatial coordinates. The larger the difference between the non-IC and the IC curves of the corresponding metrics, the better the method is capable of distinguishing random examples from real ones.}
    \label{fig:FID eval}
\end{figure*}

Another interesting observation we noticed is that the global feature-based \gls{fid} proposed by \cite{sun2020pointgrow}, to some extent, measures the "\textbf{diffusion degree}" rather than the "perceptibility" of the point clouds. For verification, we synthesize instances that are randomly distributed and therefore completely non-perceptible. We yield examples that are uniformly distributed $X_u \sim U(-r,r)$, and normally distributed $X_n \sim \mathcal{N}(0,\sigma^2)$, where $r$ increase from $0$ to $1$ and $\sigma$ grows from $0$ to $0.1$ in $10$ steps, in order to represent inputs with different "diffusion degrees".  For comparison, we stochastically choose real objects from the dataset, and calculate their \gls{fid} with objects of the same class. Theoretically, \gls{fid} performs consistently with human judgment that our randomly distributed artificial examples should exhibit significantly large \gls{fid} with real objects as they possess no recognizable geometric structures. However, as figure \ref{fig:FID eval} demonstrates, \gls{fid} (the lighter blue line) dramatically decreases with the point expansion of the instances ($r=0.1$ and $\sigma=0.02$). After the diffusion reaches the threshold ($r\approx0.2$ and $\sigma \approx0.05$), \gls{fid} fails to distinguish the meaningless point clouds from the real objects (the darker blue line), though we can still observe the discrepancies between them through \gls{cd} and \gls{emd}. A better point cloud-applicable perceptibility metric for generating examples in terms of latent distance is a promising research direction.

\subsection{AM for data reviewing} \label{sup: datareview}
\begin{figure*}
\centering
\includegraphics[width=0.6\textwidth]{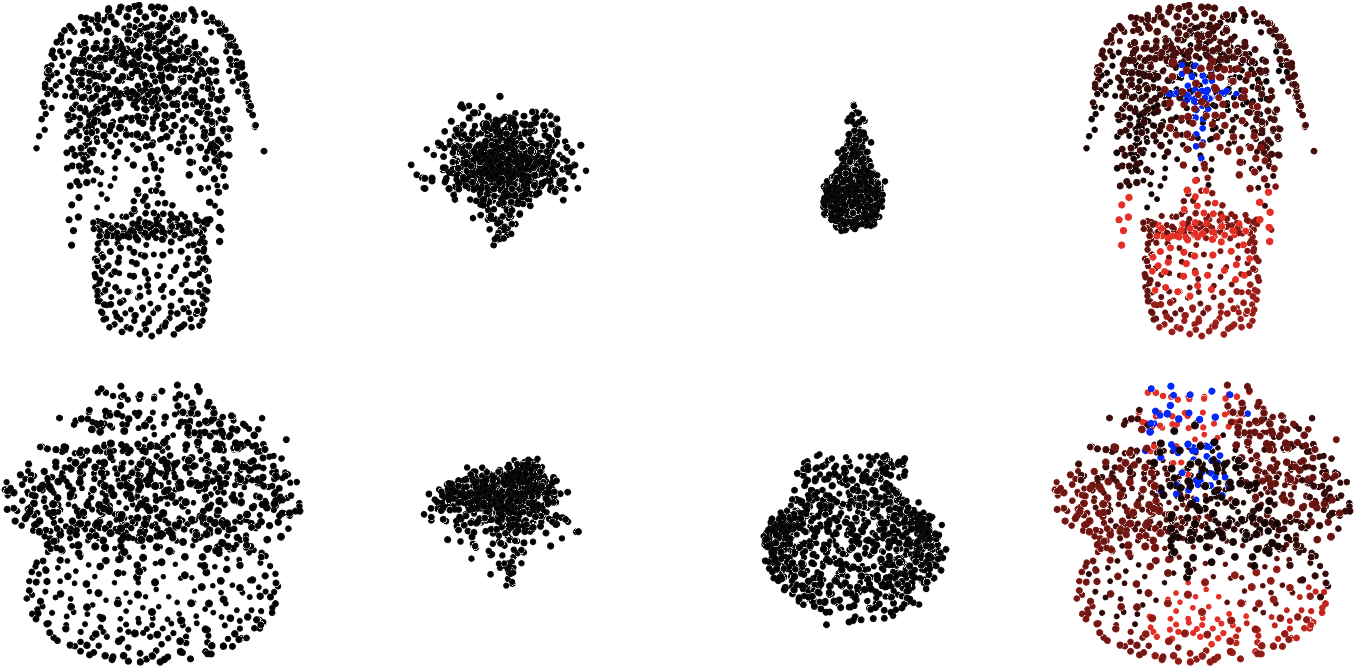}
\caption{An example of reviewing the inaccuracies of the dataset. The first column shows the instances in the dataset that are labeled as "plant" but are classified as "vase". The second and third columns demonstrate the AM output for the categories "plants" and "vases" respectively. The last column exhibits an explanation generated from 3D LIME, where brighter red points represent more positive attributions while conversely brighter blue points represent more negative attributions, neutral attributed points are colored as black.}
\label{fig_sup:AM_confusing}
\end{figure*}

Explanations can facilitate human understanding of the operating behavior of black-box neural networks. As a global explainability method, AM depicts the ideal input learned by the model. When the performance of the model is sufficiently promising, one considers that the result of AM should be a generalization of an outline of the objects from the corresponding class. Therefore, we can review those misclassified input instances utilizing this characteristic. An example is shown in figure \ref{fig_sup:AM_confusing}. Several instances in the dataset with the "plant" label are misclassified as "vase", whereas a comparison exhibits that a single "plant" label is ambiguous since the composite instance also contains the "vase" fraction. Observing the second and third columns, AM correctly describes the object outlines of the corresponding neurons in the model without any confusion. For validation, we also generate explanations for these instances employing the point cloud-applicable LIME \cite{tan2022surrogate} (the last column), the conclusions of the two explanations are approximately analogous, and the explanation given by the model is consistent with its predicted label in human perception.

%% file: 5_Conclusion.tex
\section{Conclusion} \label{conclusion}
In this work, we aim to investigate the global explanations of point cloud networks with the \gls{am} algorithm. We demonstrate that \gls{am}s based on point-wise regularizations struggle to illustrate the ideal outline of the objects, and we propose three generative model-based \gls{am} approaches which significantly enhance the perceptibility of the generated examples while also maintaining their diversity. In addition, to address the lack of \gls{am} evaluation metrics and the limitations of existing methods on point clouds, we propose a composite evaluation metric, balancing activation value, diversity and perceptibility. The results show that our generative \gls{am} methods outperform the regularization-based ones in both qualitative and quantitative aspects. For future work, we look forward to more efficient \gls{am} generation methods as well as visualizations of low-level neurons to further explore the working mechanism of point cloud neural networks. 